\pdfoutput=1
\documentclass[11pt]{article}

\usepackage{acl}
\usepackage{times}
\usepackage{latexsym}
\usepackage[T1]{fontenc}
\usepackage[utf8]{inputenc}
\usepackage{microtype}
\usepackage{graphicx}
\usepackage{booktabs}
\usepackage{tabularx}
\usepackage{multirow}
\usepackage{amsmath}
\usepackage{tikz}
\usetikzlibrary{arrows.meta,positioning,fit,backgrounds,calc}
\hypersetup{
  pdftitle={TRACE: Target-Aware Retrieval, Attributed Evidence, and Contract-Constrained Extraction for LitTraceQA},
  pdfauthor={Sachin Gupta and Divya Godara}
}
\newcolumntype{L}[1]{>{\raggedright\arraybackslash}p{#1}}
\newcolumntype{Y}{>{\raggedright\arraybackslash}X}

\newcommand{\system}{\textsc{Trace}}
\newcommand{\task}{\textsc{LitTraceQA}}

\title{TRACE: Target-Aware Retrieval, Attributed Evidence, and Contract-Constrained Extraction for LitTraceQA}

\author{Sachin Gupta \\
  Independent Researcher \\
  San Jose, USA \\
  \texttt{sachinkg12@gmail.com} \\\And
  Divya Godara \\
  Independent Researcher \\
  San Jose, USA \\
  \texttt{godaradivya@gmail.com} \\\AND
  \small\normalfont\itshape
  Accepted at the 1st Workshop on Grounding Language Models: Learning Faithfully and Efficiently \\
  (GroundLM 2026), co-located with EMNLP 2026.}

\begin{document}
\maketitle

\begin{abstract}
Finding a relevant paper is not the same as producing a verifiable answer from it.
LitTraceQA requires canonical paper identifiers, exact evidence at the page or object level, and typed answers that match the evaluator.
We call the separation between source access and scorer-visible correctness the \emph{grounding contract gap}.
\system{}---Target-Aware Retrieval, Attributed Evidence, and Contract-Constrained Extraction---addresses this gap with target-grouped retrieval, independent typed evidence localization, multimodal table extraction, schema-driven table construction, and fail-closed validation.
It indexes 27,487 papers through passage, object, alias, citation, and dense representations while retaining the question target behind each signal.
For tables, \system{} predicts the observation unit before extracting values and assembles rows with evaluator-compatible key normalization.
Our audited selected clean-track artifact scores \textbf{0.760613} on the official 71-question test set, including \textbf{0.9728} paper F1, \textbf{0.6847} evidence F1, \textbf{0.9800} multiple-choice accuracy, \textbf{0.5423} table-row F1, and \textbf{0.3508} macro cell accuracy.
On 11 public-development table records, a clean baseline and coordinate-aware visual fill obtain row F1 of 0.291 and 0.411, respectively; this diagnostic comparison includes fallback outputs and is not an official-test claim.
Remaining errors chiefly concern locator, observation-unit, row-key, and source-value identity.
\end{abstract}

\section{Introduction}

Scientific question answering is often described as retrieval followed by generation.
That abstraction is incomplete for \task{} \citep{liu2026littraceqabenchmarkmultistagegrounding}.
GroundLM 2026 pairs scientific-evidence grounding in \task{} with view-level visual-evidence identification in GoldenViewVQA \citep{wang2026doesanswercomefrom}; the shared-task findings paper describes the joint evaluation setting \citep{wang-etal-2026-groundlm}.
We participate only in \task{}, under the official evaluator team name \texttt{gabby}.
Given a research question and a fixed metadata pool, the system must produce a connected trace
\begin{equation}
q \longrightarrow \widehat{P} \longrightarrow \widehat{E}_{\widehat{P}} \longrightarrow \widehat{A},
\label{eq:trace}
\end{equation}
where $\widehat{P}$ is a set of canonical paper IDs, $\widehat{E}$ is a set of typed source locators inside those papers, and $\widehat{A}$ is one or more schema-constrained answers.
The components are scored separately, and evidence must close over the selected paper set.
A semantically correct answer can therefore score poorly when a page is adjacent to the annotated page, a printed object ID is normalized incorrectly, or a source value is placed under a paraphrased row key.

Our first pipeline had all three pathologies.
It collapsed retrieval results for several named methods into one global rank, emitted only evidence quoted by the chosen answer path, and treated every selected paper as exactly one table row.
These were not model-capacity failures.
They were representation failures at the interfaces between retrieval, grounding, and structured generation.
We call the resulting separation between information available in the corpus and information emitted in evaluator-visible form the \emph{grounding contract gap}.

We organize the system study around three questions:

\noindent\textbf{RQ1:} Which retrieval representation preserves the identity and cardinality of the requested paper targets?\\
\textbf{RQ2:} Why can source-correct evidence and values still fail structured evaluation, and which decomposition reduces that loss?\\
\textbf{RQ3:} Which execution contracts prevent operational errors from silently becoming valid but score-dead submissions?

Our contributions are:
\begin{itemize}
    \item a target-aware paper selector that retains route, target group, group-local rank, route score, and semantic role instead of flattening heterogeneous evidence;
    \item an independent typed localizer over PDF text and detected tables, figures, equations, and citation contexts, with exact scorer-key normalization and paper--evidence closure;
    \item an observation-unit-first table pipeline that plans the row axis, extracts visual and textual values, and assembles typed rows deterministically; and
    \item an artifact-level audit protocol covering corpus/index alignment, semantic fallbacks, official validation, hashes, source revision, and recorded selected artifacts.
\end{itemize}
Code, configurations, and release-audit materials are available at
\url{https://github.com/sachinkg12/trace}.

\begin{figure*}[t]
  \centering
  \includegraphics[width=\textwidth]{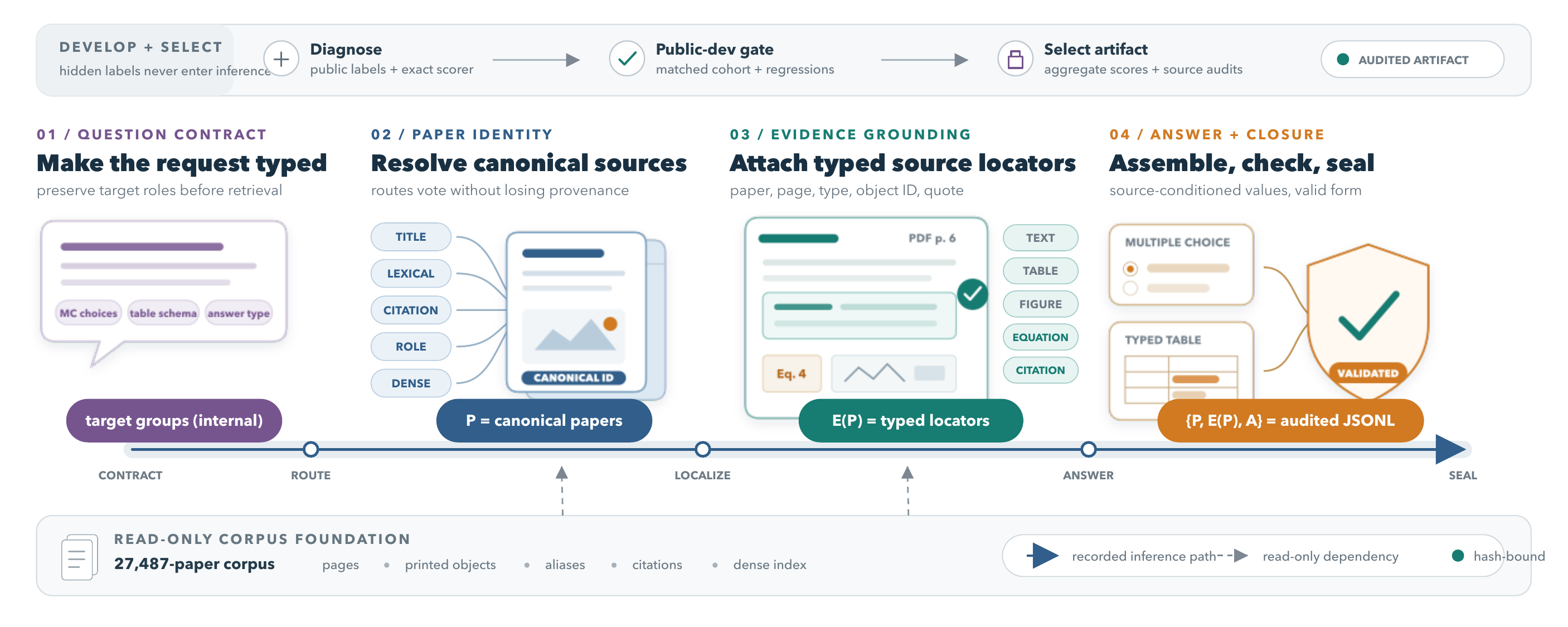}
  \caption{\system{} carries a typed request through an internal target-group contract to scorer-visible canonical papers $P$, evidence locators $E(P)$, and answer $A$. Input badges denote optional multiple-choice choices, an optional table schema, and the required answer type. A detached development-and-selection rail records public-development gates, aggregate-score checkpoint selection, source audits, and hashes. Inference uses released inputs and the corpus; hidden labels and per-question evaluator feedback are unavailable.}
  \label{fig:architecture}
\end{figure*}

\section{Task, Data, and Related Work}

\subsection{The three-part output contract}

LitTraceQA evaluates the complete path in Equation~\ref{eq:trace}, rather than treating citations as optional explanations after answering \citep{liu2026littraceqabenchmarkmultistagegrounding}.
Its evidence types include text spans, tables, figures, equations or algorithms, and citation contexts.
Paper and evidence sets receive macro precision, recall, and F1; multiple-choice answers receive accuracy; and structured tables receive row F1 plus macro and micro cell accuracy.

Our test input has 71 records: 50 request a multiple-choice component and 21 request a table component; no test record requests both.
The candidate pool contains 27,487 papers.
This scale is modest for first-stage retrieval but difficult for exact trace construction: the same question can name answer targets, evidence anchors, datasets, venues, and comparison criteria, only some of which denote papers that should be returned.

\subsection{Corpus and indexes}

We retain the organizer paper identifier as the primary key and store released PDFs in Google Cloud Storage (GCS).
PyMuPDF \citep{pymupdf2026} extracts page text and structured blocks and renders pages for visual analysis.
The frozen corpus build contains 1,409,382 passages, 500,928 detected objects, 3,711,590 aliases, and 492,892 citation/relation records.
A 384-dimensional \texttt{BAAI/bge-small-en-v1.5} \citep{xiao2023cpack} cache embeds every title--abstract pair.
Pool order, record counts, model identity, and file hashes are checked as invariants before inference.

\subsection{Relation to prior systems}

Our retrieval stage combines sparse and dense signals in the spirit of retrieval-augmented generation \citep{lewis2020retrieval}, BM25 \citep{robertson2009probabilistic}, and rank fusion \citep{cormack2009rrf}; our main finding is that fusion must follow, not erase, target grouping.
For table construction, prior work shows that scientific cell identity depends on surrounding table and paper context \citep{lou2023s2abel}, and that schema-constrained extraction can outperform unconstrained generation across heterogeneous tables \citep{bai2024schema}.
ScheMatiQ makes the observation unit explicit before discovering and populating a question-specific schema \citep{leibovitch2026schematiq}.
We adapt this principle to a fixed evaluator schema: the row entity is inferred from the question and row-key columns, while all accepted rows and locators must remain compatible with the shared-task contract.

\section{System Architecture}

Figure~\ref{fig:architecture} separates development and checkpoint selection from the recorded inference path.
The system is not a monolithic agent: generative components propose typed intermediate objects, and deterministic components enforce identity, cardinality, and serialization contracts.

\subsection{Question plan and retrieval signals}

Stage 2 uses a temperature-zero planner to extract venue/year constraints, required modalities, multiplicity, desired paper count, named anchors, and target groups.
Each target receives a semantic role: \emph{answer target}, \emph{evidence anchor}, or \emph{constraint}.
This prevents, for example, a dataset mentioned in an experimental criterion from being returned as though it were a requested paper.
Invalid planner output produces a visible conservative fallback rather than aborting or silently dropping the record.

Stage 3 queries five complementary route families: (i) name, alias, and title resolution; (ii) BM25 property and target-property search over passages; (iii) citation-context search; (iv) role-aware ownership signals; and (v) dense title--abstract retrieval.
Each candidate carries supporting snippets and route signals $(r,g,k,s,\rho)$: route $r$, target group $g$, group-local rank $k$, optional route score $s$, and role $\rho$.
Detected-object records are available to later localization and diagnostics, but the selected run did not activate the object index as a paper-retrieval route.

The group-local rank is essential.
When results for four named methods are concatenated, global rank 35 may be rank 1 for the fourth method.
Flat reciprocal-rank fusion sees weak evidence; the grouped representation sees strong coverage of a distinct requested target.
Stage 4 first applies answer-scoped venue/year constraints and then covers target groups before spending remaining slots on evidence-bearing relevance.
Explicit multiplicity determines the desired count, while a hard cap of five bounds PDF and vision work.

\subsection{Independent evidence localization}

Stage 5 fetches each selected PDF and creates a common source representation.
Text spans keep page numbers; visual objects keep printed identifiers; references keep citation IDs and contexts.
We normalize forms such as ``Eq.~(6)'' and ``Reference~[24]'' while rejecting arbitrary prose numbers as object IDs.

At stage 6, evidence is retrieved independently of the answerer.
Gemini~2.5~Flash examines bounded PDF text to propose typed locators.
Gemini~2.5~Pro vision \citep{gemini2025report} is used downstream for visual table and figure answer extraction, not for evidence localization.
Candidates are ranked by structural validity, localizer confidence, BM25 relevance, and stable source order.
They are deduplicated under the evaluator's coarse key---paper, source type, page or section, and normalized object ID---before a cap of five is applied.
This replaces answer-conditioned evidence, which had high apparent precision but omitted valid support not quoted by the final answer path.

\subsection{Observation-unit-first tables}

The table schema is not merely an output template; it defines what one row means.
Stage 7 classifies the observation unit from every column marked as a row key.
\emph{Paper Title} indicates a paper axis; \emph{Method}, \emph{Benchmark}, \emph{Author}, \emph{Reference}, or a multi-key schema indicates an entity axis.
When a question lists the desired entities, the planner constructs a closed row ledger from their shortest literal names.

Extraction then runs per source, not per output row.
Visual pages and non-visual text, equation, and citation paths propose compatible rows and typed value cells.
The assembler groups by the same normalized row-key tuple as the scorer, merges complementary cells across papers, removes all-null records, and deduplicates.
For the selected clean track, deterministic row-key matching preserves non-null baseline cells and permits only source-scoped, schema-compatible fills.
Repeated visual proposals are accepted only when they agree; otherwise the baseline table is retained.

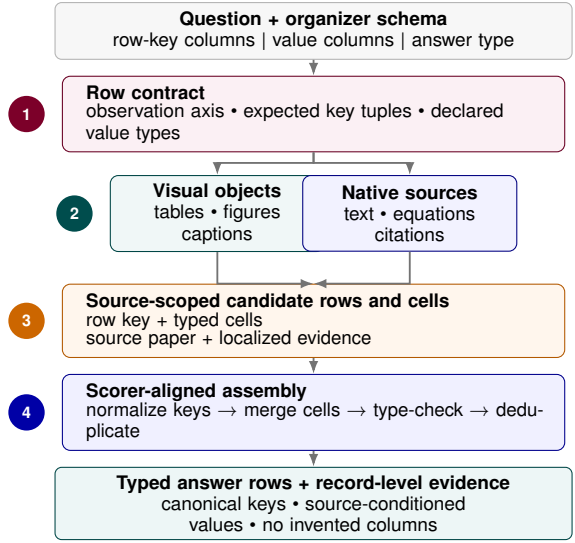
\begin{figure}[t]
\centering
\begin{tikzpicture}[
  font=\sffamily\scriptsize,
  flow/.style={-{Latex[length=1.7mm]}, line width=0.7pt, draw=black!55},
  input/.style={draw=black!30, rounded corners=2.5pt, fill=black!3,
    align=center, text width=0.86\columnwidth, minimum height=7.5mm, inner sep=3pt},
  stage/.style={draw, rounded corners=3pt, align=left,
    text width=0.78\columnwidth, minimum height=9mm, inner xsep=9pt, inner ysep=3.5pt},
  source/.style={draw, rounded corners=3pt, align=center,
    text width=0.34\columnwidth, minimum height=10mm, inner sep=3pt},
  badge/.style={circle, minimum size=4.8mm, inner sep=0pt, font=\sffamily\tiny\bfseries,
    text=white}
]
\node[input] (input) {\textbf{Question + organizer schema}\\
  row-key columns \textbar{} value columns \textbar{} answer type};

\node[stage, below=2.1mm of input, fill=purple!6, draw=purple!55!black] (contract)
  {\textbf{Row contract}\\[-1pt]
   observation axis \textbullet{} expected key tuples \textbullet{} declared value types};
\node[badge, fill=purple!65!black, left=2.2mm of contract.west] {1};

\node[source, below left=3.0mm and -1.5mm of contract.south,
  anchor=north east, fill=teal!6, draw=teal!55!black] (visual)
  {\textbf{Visual objects}\\tables \textbullet{} figures\\captions};
\node[source, below right=3.0mm and -1.5mm of contract.south,
  anchor=north west, fill=blue!5, draw=blue!50!black] (native)
  {\textbf{Native sources}\\text \textbullet{} equations\\citations};
\node[badge, fill=teal!60!black, left=2.2mm of visual.west] {2};

\node[stage, below=4.2mm of $(visual.south)!0.5!(native.south)$,
  fill=orange!7, draw=orange!70!black] (facts)
  {\textbf{Source-scoped candidate rows and cells}\\[-1pt]
   row key + typed cells\\[-1pt]
   source paper + localized evidence};
\node[badge, fill=orange!80!black, left=2.2mm of facts.west] {3};

\node[stage, below=2.1mm of facts, fill=blue!6, draw=blue!55!black] (assemble)
  {\textbf{Scorer-aligned assembly}\\[-1pt]
   normalize keys \textrightarrow{} merge cells \textrightarrow{} type-check \textrightarrow{} deduplicate};
\node[badge, fill=blue!65!black, left=2.2mm of assemble.west] {4};

\node[input, below=2.1mm of assemble, fill=teal!7, draw=teal!55!black] (output)
  {\textbf{Typed answer rows + record-level evidence}\\
   canonical keys \textbullet{} source-conditioned values \textbullet{} no invented columns};

\draw[flow] (input) -- (contract);
\draw[flow] (contract.south) -- ++(0,-1.4mm) -| (visual.north);
\draw[flow] (contract.south) -- ++(0,-1.4mm) -| (native.north);
\draw[flow] (visual.south) |- (facts.north);
\draw[flow] (native.south) |- (facts.north);
\draw[flow] (facts) -- (assemble);
\draw[flow] (assemble) -- (output);
\end{tikzpicture}
\caption{Observation-unit-first table construction. Schema keys define the row contract; visual and native sources propose source-scoped rows and cells; deterministic assembly normalizes keys, merges compatible cells, type-checks, and deduplicates. Exact PDF attestation is additionally required for any post-generation correction. The paper is a source container, not necessarily the row entity.}
\label{fig:table-pipeline}
\end{figure}

\paragraph{Public-development example.}
One multi-paper question requests four named methods (TCM, sCT, ECM-XL, and IMM) under an experimental criterion.
The legacy generator created one row per paper, confusing the paper containing a comparison with the methods being compared.
The planned system instead fixes the four methods as candidate row identities, searches each method--criterion tuple, and accepts only rows whose keys normalize to the requested set.
The example illustrates the broader distinction between \emph{source container} and \emph{observation unit}.

\subsection{Fail-closed execution}

Stages 1 and 8 make operational validity part of the architecture.
Input normalization accepts object- and list-shaped multiple-choice options while preserving organizer labels and exact table schemas.
Strict preflight checks 71 unique query IDs, 27,487 pool IDs, answer-type counts, all four multiple-choice labels, index record and byte counts, dense-cache shape/model, and official file hashes.
The runner records timeouts, semantic fallbacks, repaired components, active configuration, and raw traces.
It writes atomically only after paper--evidence closure and the pinned organizer validator pass.

Table~\ref{tab:failures} summarizes the design as recovered structure rather than a list of prompts.

\begin{table*}[t]
\centering
\small
\setlength{\tabcolsep}{5pt}
\begin{tabularx}{\textwidth}{L{0.19\textwidth}L{0.22\textwidth}Y L{0.20\textwidth}}
\toprule
Observed failure & Information lost & Architectural response & Enforced invariant \\
\midrule
Flat fusion misses later named methods & Target identity and group-local rank & Route signals preserve $(r,g,k,s,\rho)$; selection covers groups & Selected papers retain contributing route signals and target-group metadata \\
Answer-coupled evidence under-recalls & Support not quoted by the answer path & Independent typed PDF-text localization; visual extraction remains separate & Every locator closes over an emitted paper and a valid source type \\
One row per paper collapses methods/components & Observation unit and row cardinality & Row-axis classification, row ledger, multi-row extraction & Row keys and cell types conform to the supplied schema \\
Plausible values miss exact scoring & Printed form, page, or object identity & Source-conditioned candidates; PDF-exact attestation for accepted corrections & Normalization and assembly use scorer-parity keys; corrections are source-bound \\
Valid JSON hides empty behavior & Semantic success and corpus readiness & Strict preflight, fallback telemetry, atomic output & Counts/hashes pass; repairs and placeholders remain observable \\
\bottomrule
\end{tabularx}
\caption{The main failure modes and the structure restored by \system{}. The common pattern is that a downstream model cannot reconstruct information already discarded at an upstream interface.}
\label{tab:failures}
\end{table*}

\section{Experimental Setup}

\paragraph{Models and infrastructure.}
The hybrid pipeline uses Python~3.11, PyMuPDF, local BM25 indexes, and sentence-transformer embeddings.
Index construction, retrieval scoring for a fixed plan, normalization, composition, and validation are deterministic; hosted generative proposals are not.
Gemini~2.5~Flash performs question planning, text localization, and answer generation; Gemini~2.5~Pro is reserved for visual table and figure extraction.
The corpus/index store is on GCS, and bounded workers with per-record deadlines run on a Google Cloud VM.
Temperature is zero where supported, and raw replies are retained because hosted services can still vary between calls.
No task-specific model weights are trained or fine-tuned.

\paragraph{Development and freezing.}
Architecture choices and local gates use the 55-example public development split.
The table analysis below evaluates the same 11 public-development records in both conditions with organizer-compatible normalization.
Aggregate official evaluator scores informed checkpoint selection, but hidden labels and per-question evaluator feedback were unavailable and do not enter the correction policies.
The selected official artifact has 71 unique records, passes the pinned validator, and has SHA-256 prefix \texttt{2dd56b6009}.
It extends the clean 0.757968 checkpoint with three disjoint, source-attested record repairs: two table-object relocations from mention pages to their unique caption pages and one canonical method-key repair.
The terminal policy revision \texttt{a4962b9} preserves the other 68 serialized records byte-for-byte.
Each accepted delta is derived from released questions, clean full-run traces, and URL/SHA-bound PDFs; the reports bind predecessor artifacts, source revisions, runtime archives, PDF manifests, the pinned validator, and the final digest.
Projected and unscored variants are excluded from reported results.

\section{Results}

\subsection{Official result and score anatomy}

\begin{figure}[t]
\centering
\begin{tikzpicture}[font=\footnotesize, x=4.0cm]
  \def\barh{0.25}
  \draw[black!35] (0,0.50) -- (1,0.50);
  \foreach \x/\lab in {0/0,.25/.25,.5/.50,.75/.75,1/1.0} {
    \draw[black!35] (\x,0.45) -- (\x,0.55);
    \node[below, font=\scriptsize] at (\x,0.43) {\lab};
  }

  \node[anchor=east] at (-0.025,-0.15) {Paper F1};
  \fill[blue!55!black] (0,-0.27) rectangle (0.9728,-0.03);
  \node[anchor=east, font=\scriptsize\bfseries, text=white] at (0.965,-0.15) {0.9728};

  \node[anchor=east] at (-0.025,-0.55) {Evidence F1};
  \fill[green!48!black] (0,-0.67) rectangle (0.6847,-0.43);
  \node[anchor=west, font=\scriptsize\bfseries] at (0.6847,-0.55) {0.6847};

  \node[anchor=east] at (-0.025,-0.95) {MC acc.};
  \fill[orange!75!black] (0,-1.07) rectangle (0.9800,-0.83);
  \node[anchor=east, font=\scriptsize\bfseries, text=white] at (0.972,-0.95) {0.9800};

  \node[anchor=east] at (-0.025,-1.35) {Row F1};
  \fill[orange!58!black] (0,-1.47) rectangle (0.5423,-1.23);
  \node[anchor=west, font=\scriptsize\bfseries] at (0.5423,-1.35) {0.5423};

  \node[anchor=east] at (-0.025,-1.75) {Cell macro};
  \fill[orange!40!black] (0,-1.87) rectangle (0.3508,-1.63);
  \node[anchor=west, font=\scriptsize\bfseries] at (0.3508,-1.75) {0.3508};

  \node[anchor=east] at (-0.025,-2.15) {Cell micro};
  \fill[orange!28!black] (0,-2.27) rectangle (0.4023,-2.03);
  \node[anchor=west, font=\scriptsize\bfseries] at (0.4023,-2.15) {0.4023};

  \node[anchor=east] at (-0.025,-2.55) {Composite};
  \fill[purple!60!black] (0,-2.67) rectangle (0.760613,-2.43);
  \node[anchor=west, font=\scriptsize\bfseries] at (0.760613,-2.55) {0.7606};
\end{tikzpicture}
\caption{Official score profile, including both table-cell accuracies. Metrics have different definitions and are shown together only to locate residual headroom, not as an additive decomposition of the composite.}
\label{fig:score-profile}
\end{figure}
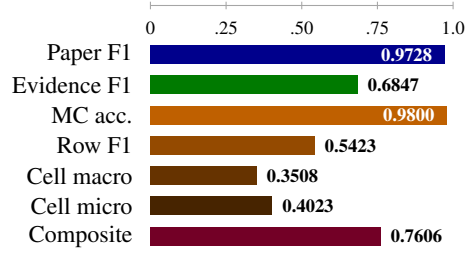

Table~\ref{tab:official} reports every metric returned for the selected clean-track official artifact.
Paper retrieval and multiple-choice answering are the strongest components, but neither is saturated.
Figure~\ref{fig:score-profile} shows the largest scorer-visible headroom in evidence and table metrics; the analysis below discusses likely identity-related causes.

\begin{table}[t]
\centering
\small
\setlength{\tabcolsep}{4pt}
\begin{tabular}{lrrr}
\toprule
Component & Precision & Recall & F1 / Acc. \\
\midrule
Papers & 0.9859 & 0.9683 & \textbf{0.9728} \\
Evidence & 0.6594 & 0.7582 & \textbf{0.6847} \\
Multiple choice & -- & -- & \textbf{0.9800} \\
Table rows & -- & -- & \textbf{0.5423} \\
Table cells (macro) & -- & -- & \textbf{0.3508} \\
Table cells (micro) & -- & -- & 0.4023 \\
\midrule
\multicolumn{3}{l}{Official composite} & \textbf{0.760613} \\
\bottomrule
\end{tabular}
\caption{Official test metrics for the selected clean-track \system{} submission. Free-form exact match was not applicable.}
\label{tab:official}
\end{table}

The selected result supersedes, but does not erase, the immediately preceding 0.757968 checkpoint or the earlier clean results at 0.739561, 0.735774, and 0.710662.
All remain in the verified clean-result history.

\subsection{Public-development table analysis}

Table~\ref{tab:table-ablation} compares a clean general table baseline with the clean framework plus coordinate-aware visual fill on the same 11 public-development table records.
The framework preserves target structure through source-container selection, then fills only schema-compatible missing cells from repeated visual agreement.
Row F1 rises by 0.120, macro cell accuracy by 0.125, and micro cell accuracy by 0.259.
An execution failure denotes a schema-valid placeholder rather than a usable table; failed records were not removed.
The baseline has three such failures and the coordinate-fill candidate has two, and the metrics score all 11 records.
Candidate evidence F1 falls from 0.460 to 0.397.
We therefore treat this as public-development diagnosis, not a promotion gate or evidence about official-test improvement.

\begin{table*}[t]
\centering
\small
\setlength{\tabcolsep}{4pt}
\begin{tabular}{lrrrrr}
\toprule
System & Paper F1 & Evidence F1 & Row F1 & Cell macro & Cell micro \\
\midrule
General clean baseline & 0.556 & \textbf{0.460} & 0.291 & 0.262 & 0.148 \\
Framework + coordinate fill & \textbf{0.594} & 0.397 & \textbf{0.411} & \textbf{0.387} & \textbf{0.407} \\
\bottomrule
\end{tabular}
\caption{Public-development diagnostics on all 11 table-answer records. Generation failures emitted schema-valid placeholder rows and remained in scoring; the baseline failed on 3/11 records and the coordinate-fill candidate on 2/11. These are not official-test metrics.}
\label{tab:table-ablation}
\end{table*}

\section{Analysis: Where Correctness Is Lost}

\paragraph{RQ1: Preserve target identity before ranking.}
Target grouping, scoped constraints, and desired cardinality solve different parts of paper selection.
The selected system's paper F1 of 0.9728 makes retrieval its strongest trace component while leaving measurable headroom.
The important negative result is flat fusion: no route weighting can recover which method produced a candidate after that group identity has been discarded.
Representation precedes ranking.

\paragraph{RQ2: Source access does not guarantee exact identity.}
Paper F1 exceeds evidence F1 (0.9728 versus 0.6847), while macro cell accuracy is 0.3508.
For evidence, an adjacent page or semantically equivalent source type is still the wrong locator tuple.
For tables, a correct number under a descriptive rather than canonical row key is still the wrong cell.
The observation-unit decomposition improves the latter, but aliases, multi-component rows, symbols, delimiters, and typed nulls remain difficult.
Our recall exceeds precision for evidence, so adding more locator candidates without stronger exact corroboration is unlikely to be the best next step.

\paragraph{RQ3: Contracts turn silent failures into actionable ones.}
Several severe bugs produced schema-valid output: empty indexes, constant-label multiple-choice fallbacks, all-null rows, and repaired answer objects.
Record counts, content hashes, closure checks, and semantic fallback telemetry distinguish these cases from genuine model errors.
The lesson is broader than this task: when a pipeline combines retrieval, hosted models, PDF rendering, and structured output, the artifact boundary must be tested as rigorously as the model.

\paragraph{What did not work.}
Unbounded multi-paper plans propagated dozens of PDFs into per-paper vision and exhausted deadlines.
Answer-conditioned evidence improved apparent precision but cut recall.
Flat reciprocal-rank fusion promoted generic high-BM25 papers over later named targets.
Unconstrained table generation paraphrased row keys and attached baseline values to the container paper.
More model calls did not repair these losses reliably; explicit intermediate identities did.
The four-record extension from 0.739561 to 0.757968 provides a sharper positive result: paper F1 rises from 0.9587 to 0.9728, evidence F1 from 0.6753 to 0.6847, row F1 from 0.4709 to 0.5185, and macro cell accuracy from 0.3032 to 0.3508, while multiple-choice accuracy remains 0.9800.
The macro table increase of 0.0476 is consistent with one $1/21$ record-level increment and comes from binding requested metric rows and columns to physical PDF coordinates before canonicalizing decorative direction glyphs.
The evidence gain comes from source-local paper correction and one uniquely re-localized answer-bearing page, rather than increasing candidates globally.
Source attestation is necessary, but table and evidence changes additionally require scorer-contract identity and a fail-closed mutation boundary.
The final three-record extension raises only row F1, from 0.5185 to 0.5423; paper, evidence, multiple-choice, and cell metrics are unchanged.
This isolates the observed gain to canonical method identity, while the two caption-page relocations are source-correct but evaluator-neutral.

\subsection{Lessons from adaptive diagnosis}

A historical adaptive artifact reached 0.792252 only after repeated official-score diagnosis and manual row/cell adjudication.
It is not a comparable system result and is excluded from our selected clean-track claim.
Its value is diagnostic: the gap showed that source access and model capacity were not the principal bottlenecks; exact paper ownership, locator identity, observation units, row keys, and cell coordinates were.
Those findings motivated the identity ledger, source-object contracts, and fail-closed mutation boundaries described here.
We therefore report the number once as a lesson about system interfaces, not as the headline performance of \system{}.

\section{Discussion and Reproducibility}

The prior 0.757968, 0.739561, 0.735774, and 0.710662 clean checkpoints remain in the result ledger rather than being relabeled as adaptive or discarded.
The selected 0.760613 artifact adds only generic, fail-closed corrections: exact table-caption localization and canonical method identity derived from immutable PDFs and the released schema.
Its composition applies complete records by query ID, rejects overlapping chains, and preserves 68 of 71 records from the preceding clean checkpoint byte-for-byte; no hidden labels or per-question evaluator feedback enter these policies.
Aggregate official scores did inform checkpoint selection, which we distinguish from the source-only inputs to each correction policy.

The public full-generation code implements and can execute the base architecture from released inputs and source assets, but hosted proposals may vary and the released configuration does not regenerate the exact selected artifact.
In particular, the selected base used \texttt{planned\_visual\_fill} table extraction, whereas the public general configuration uses \texttt{planned}; the selected correction chain is retained in the audit package rather than exposed as the public full-generation command.
An independent run of that public configuration matched 37 of 71 selected records as complete serialized objects, with most variation arising in table construction; we therefore do not claim exact output or score reproducibility.
The exact 0.760613 JSONL is uploaded separately with the paper and is reported as an audited selected artifact; our retained audit package binds predecessor reports, source revisions, PDF manifests, output digests, and validator status.

\section{Conclusion}

The audited selected clean-track \system{} artifact reaches 0.760613 by preserving scorer-visible paper, evidence, and answer identities through target-aware retrieval, independent localization, observation-unit-first rows, and fail-closed validation.

\section{Limitations}

The public-development table analysis covers only 11 records and cannot establish official-test improvement.
The fixed five-paper cap can under-recall unusually large answer sets.
PDF parsing remains brittle for scans, complex multi-column layouts, and objects whose printed identifiers are absent.
Exact evidence evaluation can penalize semantically valid adjacent support; the LitTraceQA task paper itself notes that locator normalization remains a benchmark-design challenge \citep{liu2026littraceqabenchmarkmultistagegrounding}.
Hosted model APIs introduce cost, version drift, and limited repeatability despite constrained prompts and retained traces.
We perform no weight updates, task-specific fine-tuning, synthetic training, or external factual augmentation.
The system operates only on the released scientific-paper pool, but its predictions can still misattribute claims or overstate support and should not replace inspection of the cited source.

\section*{Acknowledgments}

We thank the GroundLM 2026 organizers for organizing the shared task and maintaining its evaluation infrastructure.

\bibliography{references}

\appendix
\section{Reproducibility Checklist}

The public release includes implementation code, a general configuration, result documentation, and validation instructions; the OpenReview submission includes the ACL-formatted PDF and exact 71-line selected prediction file.
Our retained audit package additionally includes SHA-256 manifests, source revisions, resolved selected-run configuration, preprocessing/index statistics, predecessor reports, and validator records.
The historical adaptive artifact and public-development table analysis are separately labeled and excluded from the selected-system claim.
The official ACL style files are used without modifications to margins, spacing, fonts, or page dimensions.

\end{document}